\documentclass[review]{elsarticle}

\usepackage{lineno,hyperref}
\usepackage{subcaption}
\usepackage{amsfonts}
\usepackage{palatino}
\usepackage{mathpazo}
\usepackage{mathtools}
\usepackage{algorithmic}
\usepackage{mathabx}
\usepackage{soul}
\usepackage{lineno}
\usepackage{amsmath}
\usepackage{blindtext}
\usepackage{multirow}
\usepackage{array, longtable, tabularx}
\usepackage{adjustbox}
\modulolinenumbers[5]

\journal{Journal of Biomedical Signal Processing and Control}

\usepackage{pdflscape}
\usepackage{hyperref}%

\begin{document}

\begin{frontmatter}

\title{Reservoir Computing Models for Patient-Adaptable ECG Monitoring in Wearable Devices}

\author[add1,add2]{\small Fatemeh Hadaeghi}
\address[add1]{\footnotesize Institute of Computational Neuroscience, University Medical Center Hamburg-Eppendorf (UKE), 20251 Hamburg, Germany}
\address[add2]{\footnotesize Department of Electrical Engineering and Computer Science, Jacobs University gGmbH, Bremen, 28759 Bremen, Germany}

\begin{abstract}
The reservoir computing paradigm is employed to classify heartbeat anomalies online based on electrocardiogram signals. Inspired by the principles of information processing in the brain, reservoir computing provides a framework to design, train, and analyze recurrent neural networks (RNNs) for processing time-dependent information.
Due to its computational efficiency and the fact that
training amounts to a simple linear regression, this supervised
learning algorithm has been variously considered as
a strategy to implement useful computations not only on
digital computers but also on emerging unconventional hardware
platforms such as neuromorphic microchips. Here, this
biological-inspired learning framework is exploited to devise
an accurate patient-adaptive model that has the potential
to be integrated into wearable cardiac events monitoring
devices. The proposed patient-customized model was trained
and tested on ECG recordings selected from the MIT-BIH
arrhythmia database. Restrictive inclusion criteria were
used to conduct the study only on ECGs including, at least,
two classes of heartbeats with highly unequal number of
instances. The results of extensive simulations showed this
model not only provides accurate, cheap and fast patient-customized
heartbeat classifier but also circumvents the
problem of “imbalanced classes” when the readout weights
are trained using weighted ridge-regression.
\end{abstract}


\end{frontmatter}


\section{Introduction}
%
%
%
%
Patient-customized electrocardiogram (ECG) analysis is becoming increasingly important in clinical applications. The latest generation of event and Holter monitor systems, patch-type electrocardiogram recorders \cite{yan20100} and wearable devices with ECG sensors \cite{dias2018wearable} offer continuous recording and monitoring of the real-time cardiac data over time courses of days or weeks (i.e. beyond the traditional 48 hours) allowing for a detailed statistical analysis and a clearer picture of a patient's cardiovascular health. Real-time, patient-customized monitoring can also enable immediate intervention, or patient callbacks if an appropriate model has been exploited to diagnose patient-specific conditions. However, in order to manage this vast amount of data, and to extract the relevant information for conducting further statistical analyses, smart automated algorithms which are compatible with emerging hardware technology must be developed.

Over decades of research, many signal processing techniques such as frequency analysis, template-matching, wavelet transform, filter banks, and hidden Markov models as well as different variations of neural networks have been employed for arrhythmia detection or for the classification of cardiac beats \cite{hu1997patient,christov2006comparative,faezipour2010patient,mateo2016efficient}. However, due to the lack of a complete set of algorithms compatible with emerging micro-device technologies, the practical exploitation of automated ECG diagnostic systems is not yet very extensive, and further improvements are needed in the field of automatic portable ECG interpretation.

The strong variations in the temporal course and morphology of ECG waveforms of different patients and patient groups are a major problem in automated ECG monitoring tasks such as beat classification. One obvious, yet computationally expensive solution to this problem is exploiting a multitude of training data recorded from many subjects -- with different healthy or pathological cardiac conditions -- to develop a generic classifier. This is the approach typically taken by deep-learning inspired methods to provide predictive analytic solutions when data volume and complexity of the task are inordinate \cite {kiranyaz2016real,kachuee2018ecg}. However, such an approach for ECG diagnostic purposes faces several challenges, among them data collection, beat annotation, and the technical challenges associated with hardware implementations. In contrast to a generic model, a patient-customized classifier does not rely on a vast amount of data collected from other subjects and make the classification algorithm adaptable to the unique characteristic features of each patient’s ECG records.

In this study, the echo state network (ESN), an instance of reservoir computing (RC) models, is exploited to design a real-time patient-adaptable ECG beat classifier. The basic idea of reservoir computing is to use a random excitable medium to expand an input signal into a higher-dimensional (nonlinear transform) signal space. The generality of this idea has invited researchers from computing theory and microchip technologies to consider RC as a computational paradigm for use in ``unconventional'' physical or computational media that differ from neural network models or digital computing circuitry \cite{tanaka2019recent}. Since functional reservoirs have been successfully created in electrical circuits \cite{schrauwen2008compact,he2018EMBS}, optical media \cite{freiberger2017chip}, or chemical (molecular) substrates \cite{goudarzi2013dna}, it is promising to design and implement implantable or wearable biosensors, processors, and controllers based on this computational scheme. As discussed below, the model provides a predictive solution for the electrocardiogram (ECG) beat classification problem that outperforms the current state-of-the-art when it is trained and tested on ECG recordings selected from the MIT-BIH arrhythmia database \cite{goldberger2000physiobank,moody2001impact}. Applying the idea of ``reservoir transfer learning'' \cite{he2018EMBS}, this reservoir model has been proven to be compatible with the \textit{Dynap-se} processor, a neuromorphic hardware used for implementing analog spiking neural networks \cite{moradi2018scalable}. 

The paper is organized as follows. Section \ref{Related Works} presents a review of the literature and introduces the idea of reservoir computing. Data acquisition, preprocessing procedure and the proposed ESN model are discussed in Section \ref{Methods}. Section \ref{Results} reports the results of the obtained classifier on the MIT-BIH arrhythmia database and discusses the results. Conclusions are presented in Section \ref{Conclusion}.

\section{Related Work}\label{Related Works}
\subsection{ECG Beat Classification Techniques}

Decision-tree approaches based on various features extracted from each heartbeat have been extensively exploited for automated ECG beat classification. Such a classification system typically comprises three modules for heartbeat segmentation, feature extraction, and classification. In most cases, non-overlapping sliding windows with different shapes and lengths are initially used to segment cardiac beats for further processing. Then, various temporal features (e.g., the width and height of QRS complex, RR interval, QRS complex area)\cite{de2004automatic,christov2006comparative}, frequency domain descriptors (usually extracted using power spectral density (PSD), or discrete Cosine transform) \cite{dutta2010correlation,chen2017heartbeat}, time-frequency domain representatives (obtained using discrete Wavelet transform) \cite{christov2006comparative,faezipour2010patient} as well as features extracted from phase-space reconstruction of ECG records \cite{ubeyli2010recurrent,nejadgholi2011using} are used as input to train a classifier. A significant challenge associated with these features is their susceptibility to variations of ECG beat morphology and temporal characteristics not only among different patients or patient groups, but also for the same patient in various activity states. Therefore, the potential benefit of patient adaptation seems substantial and is worth further pursuing.

For the classification module, artificial neural networks (ANNs) \cite{ince2009generic,li2017high}, self-organizing maps (SOMs) \cite{hu1997patient}, support vector machines (SVMs) \cite{dutta2010correlation,chen2017heartbeat}, linear discriminators \cite{de2004automatic,de2006patient}, conditional random fields \cite{de2012weighted} and neuro-fuzzy networks \cite{engin2004ecg} have been exploited for discriminating heartbeats. 
To the best of my knowledge, \cite{hu1997patient} was the first paper in which a ``mixture-of-experts'' (MOE) approach was proposed to demonstrate the feasibility of having a patient-adaptable ECG beat classification
algorithm. Performance metrics averaged on 20 ECG signals selected from the MIT-BIH arrhythmia database were reported as 94.0\% accuracy, 82.6\% sensitivity, and 97.1\% specificity. Another patient-customized heartbeat classification scheme was later proposed in \cite{engin2004ecg} where the wavelet transforms variance, third-order cumulant and autoregressive (AR) model parameters served as features and a fuzzy-hybrid neural network including a fuzzy c-means
classifier and an MLP neural network has been trained to discriminate normal from abnormal cardiac beats. On 7 ECG recordings from the MIT-BIH arrhythmia database, 93.5\% accuracy, 99.6\% sensitivity, and 95.3\% specificity were obtained. The classifier has been trained and tested on a short segment of each ECG signal (i.e., 200 heartbeats). 
A three-class classification scheme based on cross-spectral density information extracted in the frequency domain was developed in \cite{dutta2010correlation} to detect normal beats, premature ventricular contractions (PVC) beats and other beats in two-lead ECG streams. The scheme, when employed for 40 files in the MIT-BIH arrhythmia database, was shown to result in classification accuracy in the range 95.51--96.12 \%.

In order to characterize ECG patterns specific to a patient, \cite{faezipour2010patient} suggested to consider the ECG waveforms as data-packet streams and apply packet-processing techniques. The method relies on precise localization of fiducial ECG points and exploits wavelet analysis with adaptive thresholding for ECG preprocessing and feature extraction. Aiming at abnormal beat detection, a classification accuracy of 97.42\% was reported on all the 47 files in the MIT-BIH arrhythmia database. 

\subsection{Reservoir Computing}
Inspired by the brain’s ability to process information,
reservoir computing provides a framework to design, train and analyze recurrent neural networks (RNNs) for processing time-dependent information \cite{jaeger2001echo,maass2002real}.

\begin{figure}
\centering
\includegraphics[width= 0.7 \linewidth]{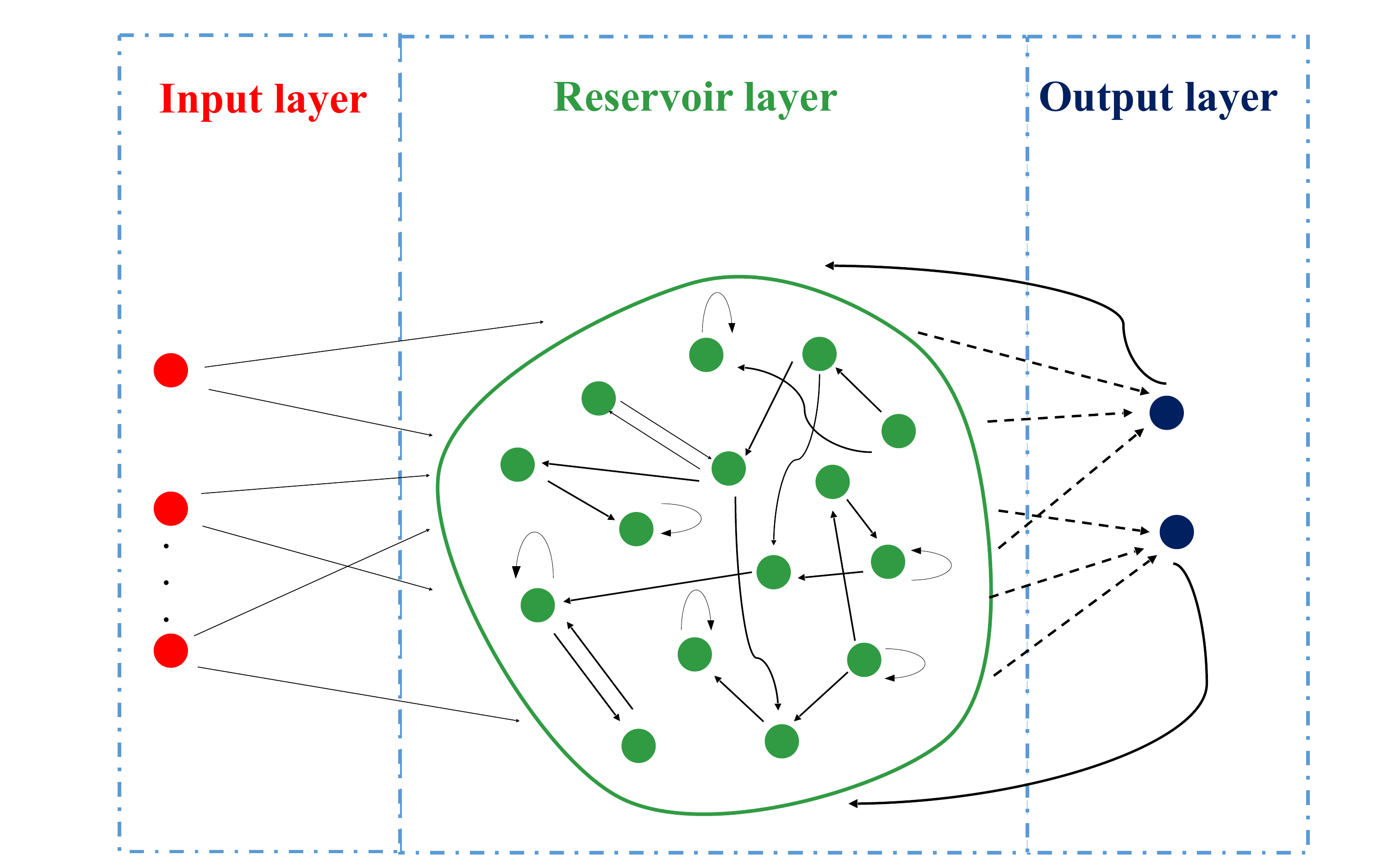}
\caption{A schematic of reservoir computing.}
\label{fig_rc}
\end{figure}

A reservoir computer comprises three major parts 
(Fig. \ref{fig_rc}). The input layer feeds the input signal into a
random, large, fixed recurrent neural network that constitutes
the ``reservoir''. The input signal is non-linearly
mapped into a higher dimensional signal space through the internal
variables of this dynamical system (i.e., reservoir
states). In the output layer, a linear combination of the
reservoir states is computed as the time-dependent output
of the reservoir. Depending on the task, randomly
generated output to reservoir (all-to-all) feedback connections
may also be included in the architecture. In contrast to traditional (and ``deep'') RNN training methods, the RC
technique proposes to only adapt the output weights
in order to minimize the mean square error between
the target and the output signal. The values of input
weights and reservoir connection weights are not critical
and can be selected at random within some pre-defined
intervals to obtain the best performance. Regarding
the fact that only the output connections are trained
and the optimization of the output layer only consists
in a linear regression, training algorithms are computationally efficient and straightforward.

The nonlinear expansion of the input signal into a
high-dimensional (reservoir) signal space, plus ease of
training enable reservoir computers to efficiently accomplish
a large range of complex tasks on time-dependent
signals. Nonlinear channel equalization \cite{jaeger2004harnessing},
time series prediction \cite{jaeger2001echo,li2012chaotic}, speech recognition \cite{triefenbach2010phoneme} and robot control \cite{antonelo2008event} are some of the examples
in which reservoir computing has been reported to
perform well.

\section{Materials and Methods}\label{Methods}
\subsection{ECG signals}

Our study was conducted on 17 ECG recordings from the MIT-BIH
arrhythmia database where each recording provides 30-minutes excerpts of two-channel (i.e., limb lead II and one of the modified leads V1, V2, V4 or V5) ambulatory ECG signals \cite{goldberger2000physiobank,moody2001impact}. Restrictive inclusion criteria were applied to select only 17 (out of 47) ECG signals including at least two classes of cardiac events with more than 50 occurrences in 30-minutes. The recordings were digitized at 360 samples per second with 11-bit resolution over a 10 mV range. Every single QRS from each recording has been independently annotated by two or more cardiologists.

Since only QRS fiducial points are annotated in the database, in our study, the entire time course of each heartbeat was initially identified by sliding overlapping rectangular windows on discrete signal of annotated QRS points. Numerous cardiac abnormalities, however, manifest themselves in changes in temporal characteristics of ECG signals such as the duration of each heartbeat and its corresponding beat intervals. For instance, in pathological conditions, P-wave duration, QRS span, and S-T interval, are either shorter or longer than in normal cardiac beats. Hence, rectangular windows with different lengths were applied to segment and label the entire course of different classes of heartbeats to provide the ground truth for the proposed supervised learning scheme. 15-minute ECG signals from lead II of 17 recordings were selected to conduct this study only on ECGs including at least two classes of heartbeats with a highly unequal number of instances. Signal drift and power-line interference were suppressed by applying a third-order band-pass Butterworth filter with cut-off frequencies of 0.4 and 45 Hz. By extracting 15-minute ECG signals corrupted by noise and artifacts, it was sought to challenge the performance of reservoir computing method dealing with the real-world data.

\subsection{Beat Classification}
We initially design an appropriate reservoir architecture which is able to detect all the annotated QRS complexes in three randomly selected ECG recordings. Based on the states of this reservoir's units in response to the input ECG, a patient-customized readout layer is then trained for the classification of different classes of cardiac beats. 

The dynamics of a typical reservoir computer with real-time continuous value units is governed by the following equations:

\begin{equation}\label{eq1}
	 \mathbf{x}(t+\Delta t) = f (\mathbf{W}\mathbf{x}(t)+\mathbf{W}^{in}\mathbf{u}(t+\Delta t)+\mathbf{W}^{fb}\mathbf{y}(t)),
\end{equation}
where $\Delta t$ is the real time sampling period, $\mathbf{u}(t) \in \mathbb{R}^{N_{u}}$ is the input signal, $\mathbf{x}(t) \in \mathbb{R}^{N_{x}}$ is the $N_{x}$-dimensional reservoir state, $f$ is a nonlinear function (usually the logistic sigmoid or the tanh function in software implementations). $\mathbf{W}$, $\mathbf{W}^{in}$ and $\mathbf{W}^{fb}$ are the input, recurrent and output feedback weight matrices, respectively. The output, $\mathbf{y}(t)\in \mathbf{R}^{N_{y}}$ is then obtained from the extended system state (i.e., $ \mathbf{z}(t) = [\mathbf{x}(t); \mathbf{u}(t)] $ where $[.;.]$ stands for a vertical vector concatenation):

\begin{equation}\label{eq2}
\mathbf{y}(t) = g(\mathbf{W}^{out}\mathbf{z}(t)),
\end{equation}
where $g$ is an output activation function (typically the identity or a sigmoid) and $\mathbf{W}^{out}$ is the readout weight matrix.

In this study, the leaky integrator (LI) neuron model was adopted whose leakage rate can be leveraged to control the ``speed'' of the reservoir dynamics. LI neuron typically performs a leaky integration of its activation from previous time steps:

\begin{equation}\label{eq3}
\begin{aligned}
& \mathbf{x}(t+\Delta t) = \\
& (1-\alpha)\mathbf{x}(t) + \alpha f (\mathbf{W}\mathbf{x}(t) + \mathbf{W}^{in}\mathbf{u}(t+\Delta t) + \mathbf{W}^{fb}\mathbf{y}(n).
\end{aligned}
\end{equation} 

By changing the leakage parameter, $\alpha \in [0, 1]$, it is possible to adjust the speed at which the reservoir reacts to the input and to tune the effective interval of frequencies in which the reservoir functions. Higher values of the leaky coefficient result in reservoirs that react rapidly to the input \cite{jaeger2007optimization}.

\begin{figure*}
    \centering
    \begin{subfigure}[t!]{0.3\textwidth}
        \includegraphics[width=\textwidth]{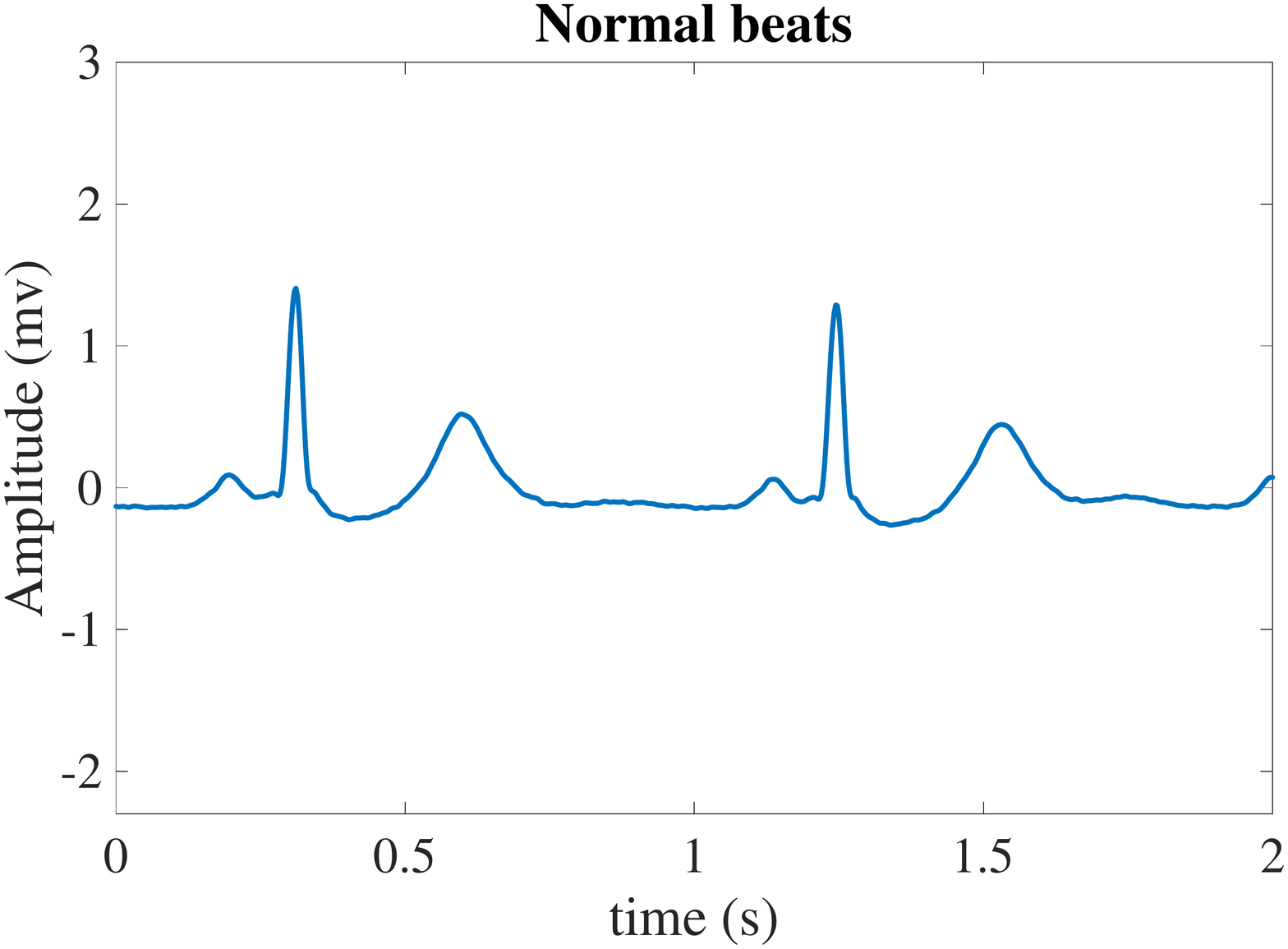}
        \includegraphics[width=\textwidth]{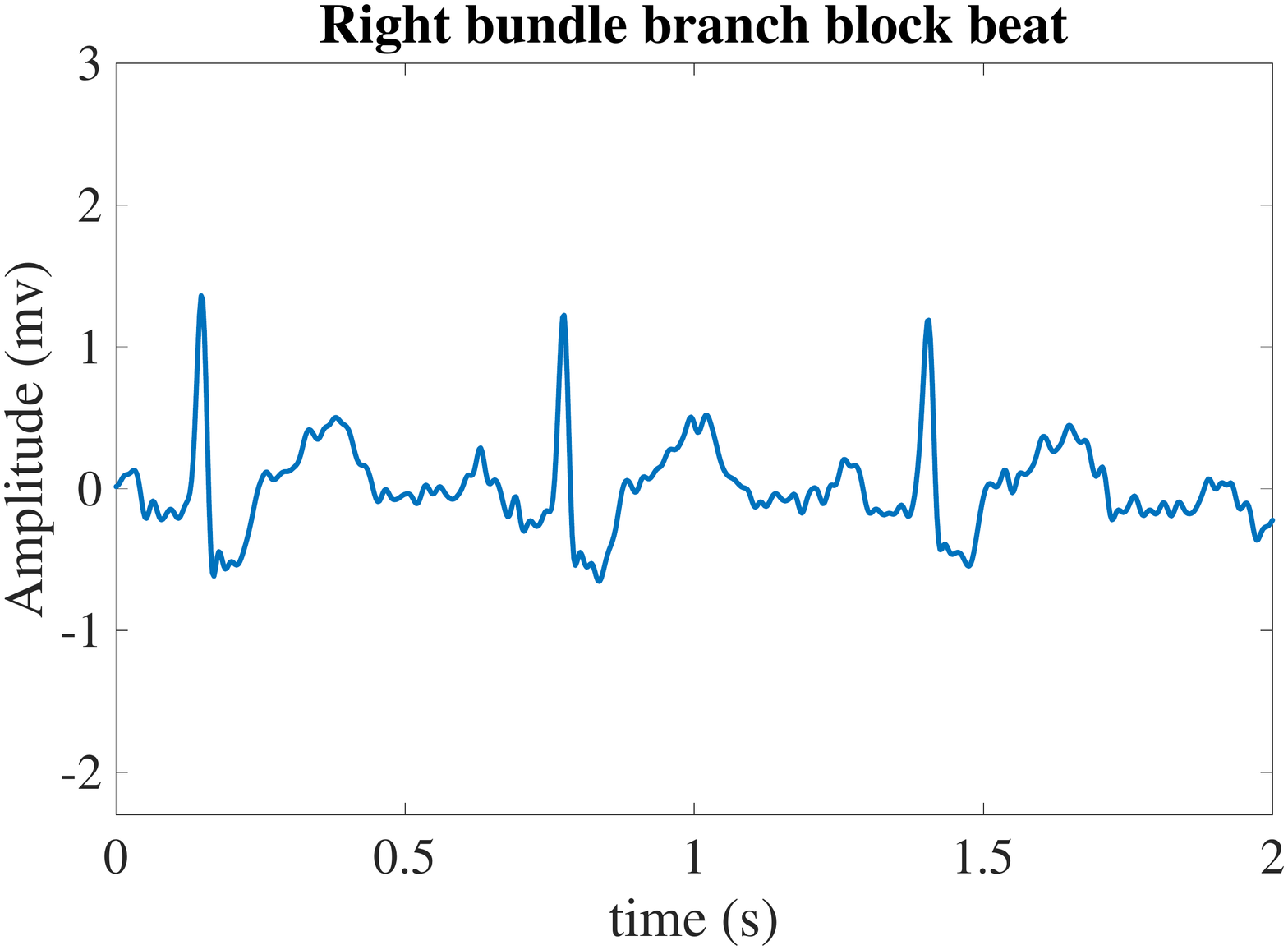}
        \includegraphics[width=\textwidth]{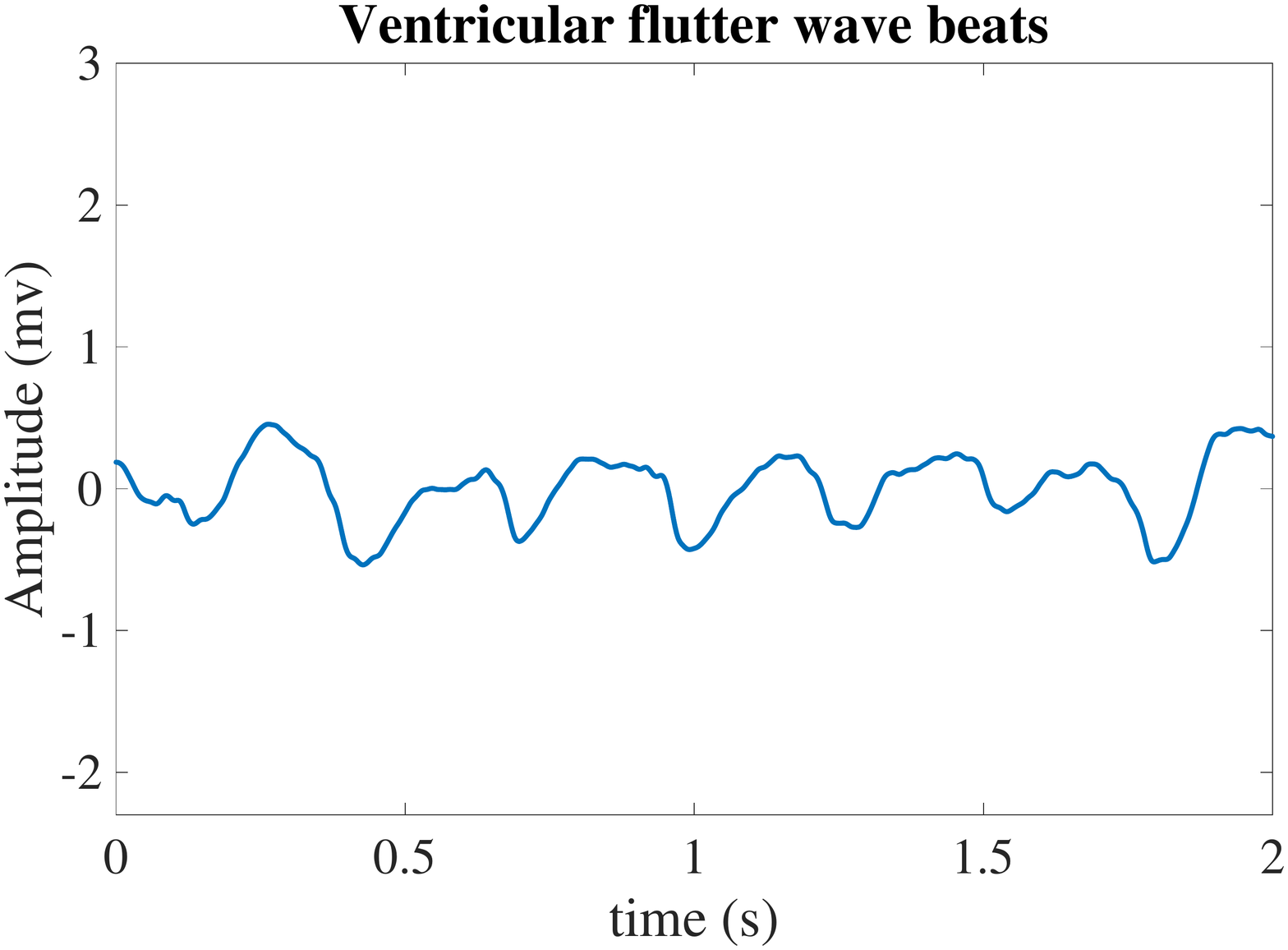}
        \end{subfigure}
    \centering
    \begin{subfigure}[t!]{0.3\textwidth}
    \includegraphics[width=\textwidth]{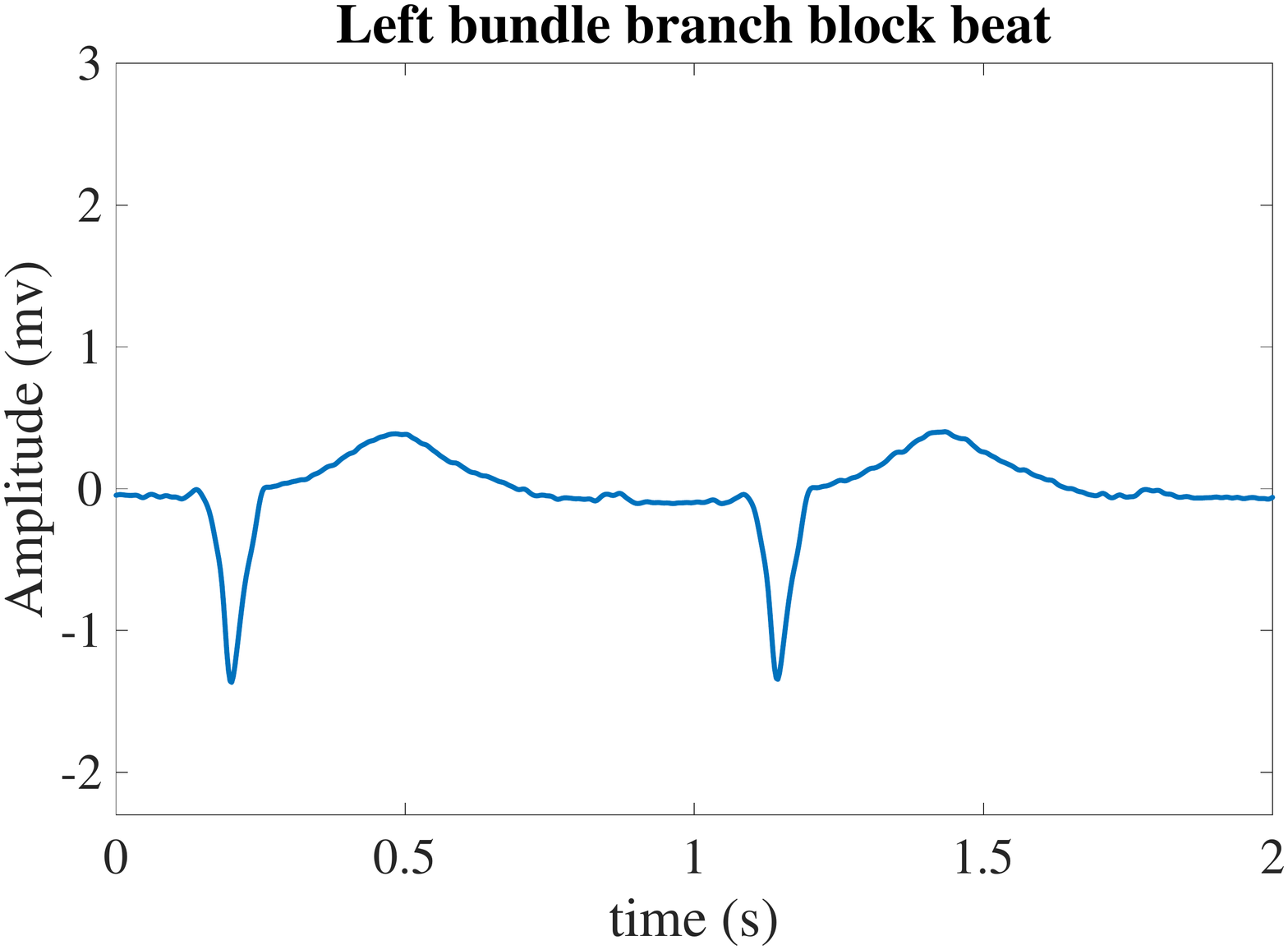}
    \includegraphics[width=\textwidth]{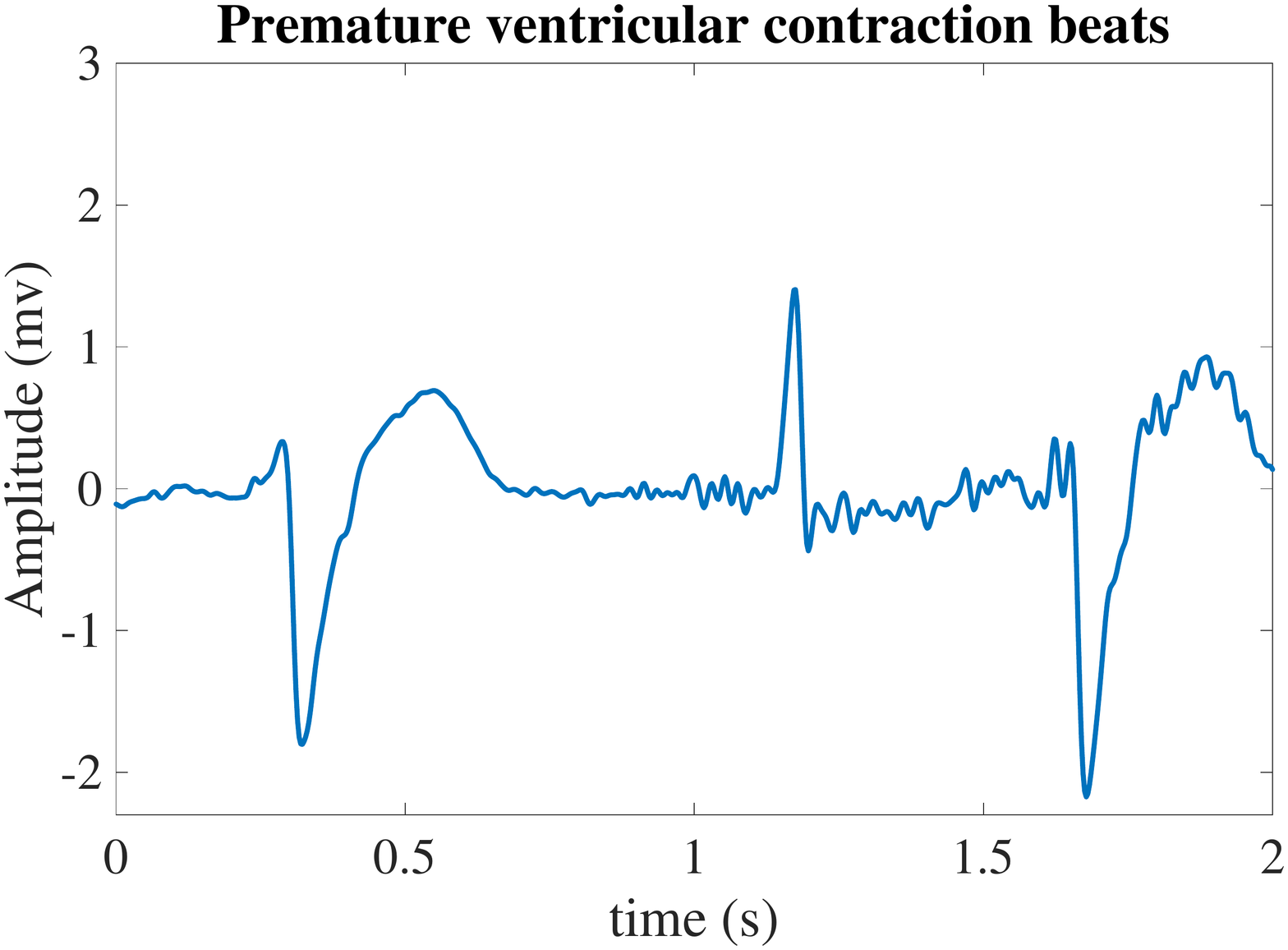}
    \includegraphics[width=\textwidth]{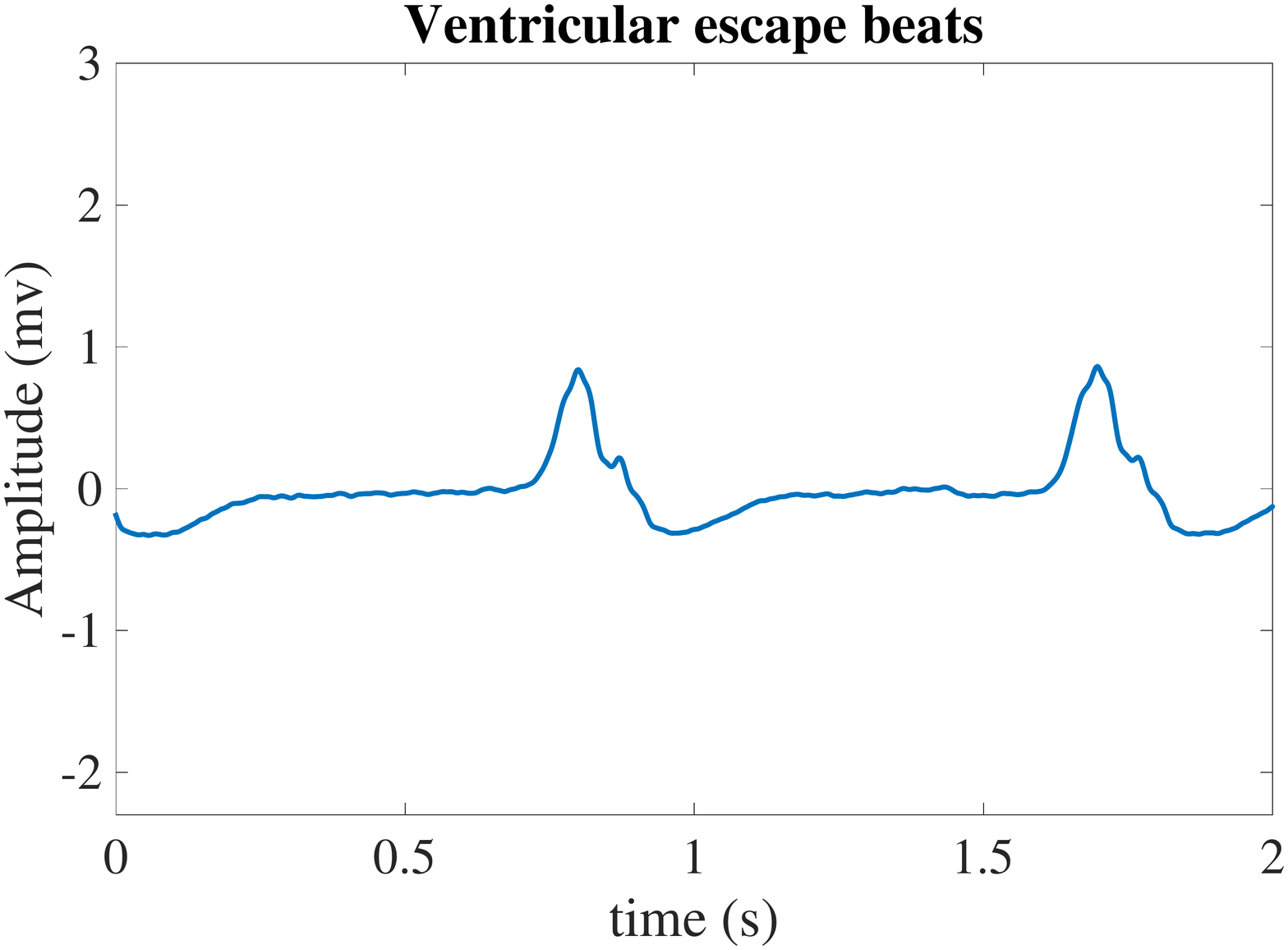}
   
    \end{subfigure}
    \centering
    \begin{subfigure}[t!]{0.3\textwidth}
        \includegraphics[width=\textwidth]{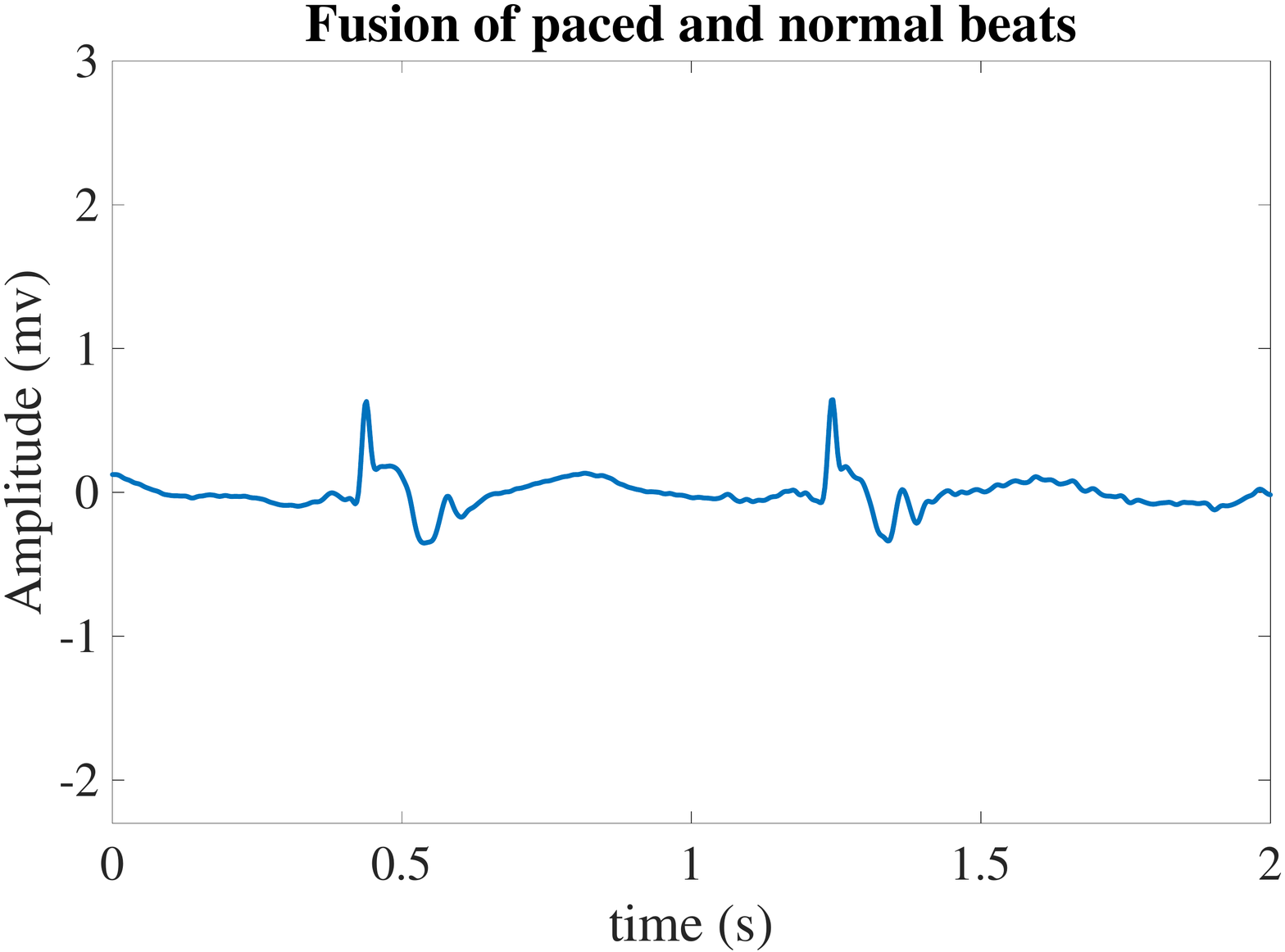}
        \includegraphics[width=\textwidth]{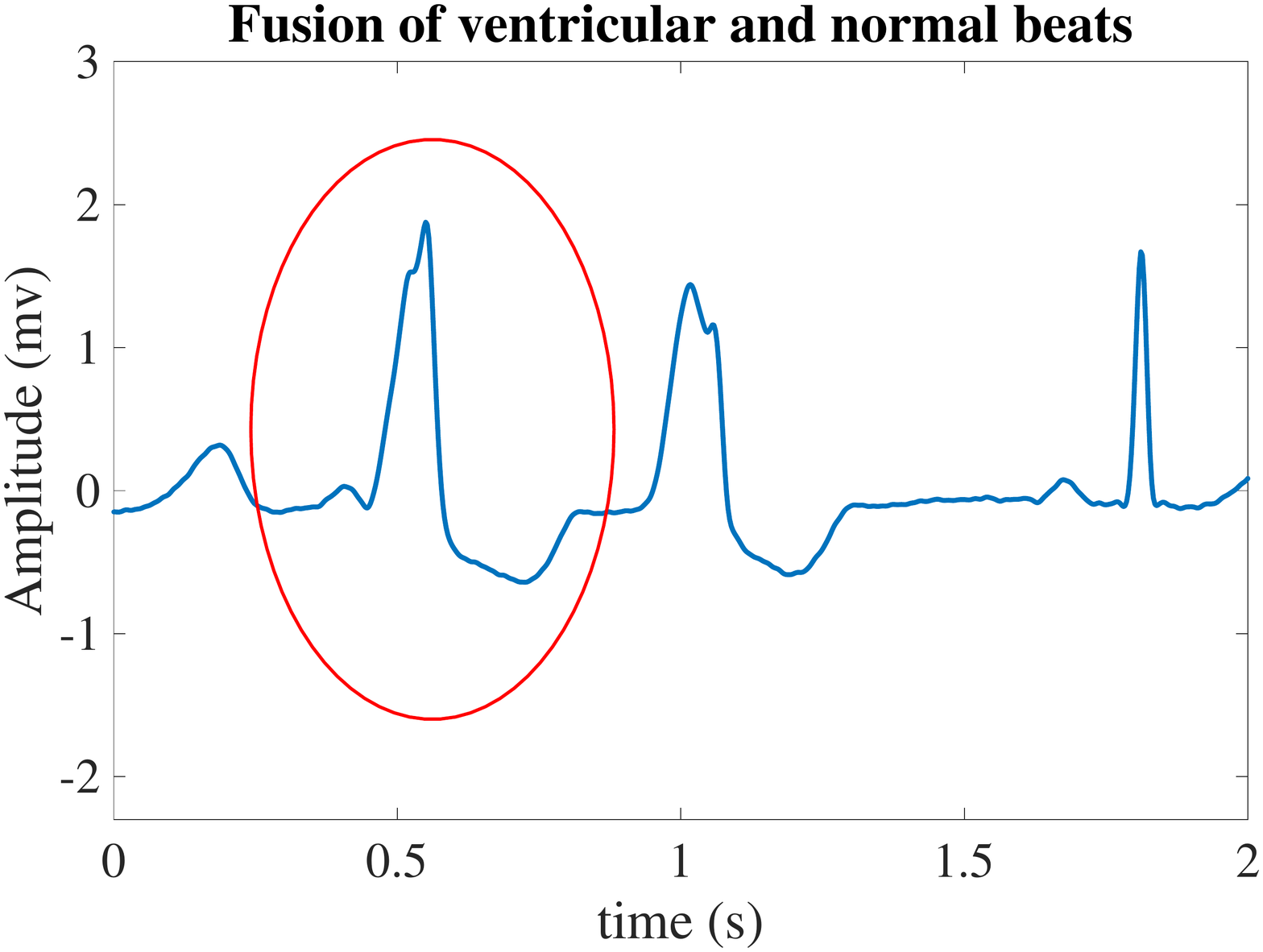}
         \includegraphics[width=\textwidth]{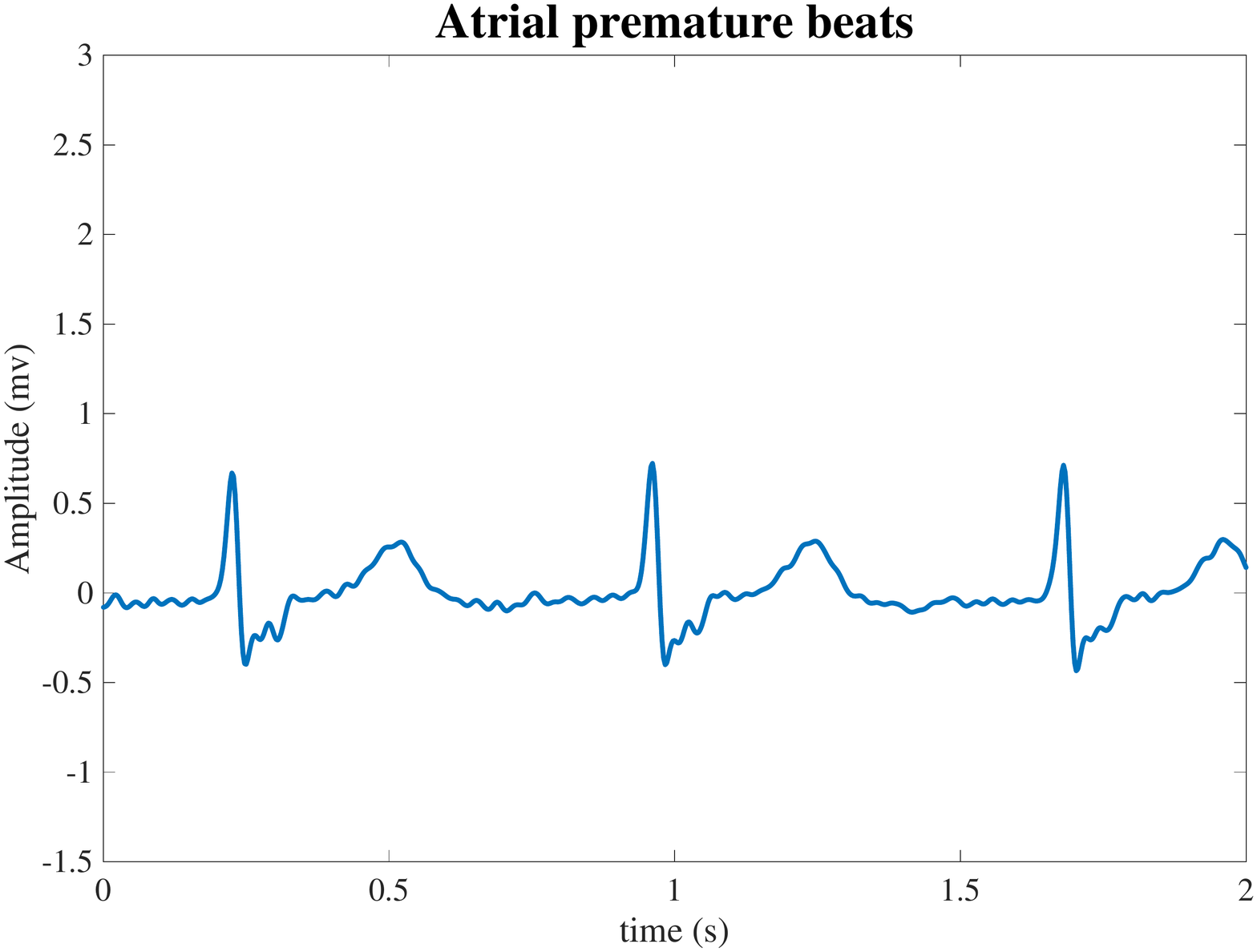}
    \end{subfigure}
    \caption{Schematic representation of normal and pathological cardiac beats.}
    \label{beatSample}
\end{figure*}

Training the readout is typically done by computing the linear regression weights of the target outputs on the harvested states of reservoir units and the inputs via the pseudo-inverse method, the Wiener-Hopf solution, or ridge-regression \cite{jaeger2001echo}. With $N_{y}$ outputs, $N_{x}$ reservoir units and $N_{u}$ inputs, given the set $ \{y_{1i} , ..., y_{li}, z_{1i} , ..., z_{ki}\}_{k=N_{x}+N_{u}, l = N_{y}, i=1,...,N} $ of $N$ observations, ordinary least square (OLS) regression model is commonly used in standard RC to define the regression coefficients:

\begin{equation}\label{eq4}
	 \mathbf{W}^{out} = \mathbf{Y}_{target}\mathbf{Z}^{T}(\mathbf{Z}\mathbf{Z}^{T} + \gamma^{2}I)^{-1}
\end{equation}
where $\mathbf{Z} \in R^{(N_{x}+N_{u}) \times N}$ is the matrix of extended system states, $\mathbf{Y}_{target} \in R^{N_{y} \times N}$ contains the time courses of target outputs, $I \in R^{(N_{x}+N_{u}) \times (N_{x}+N_{u})}$ is the identity matrix and $\gamma \geq 0$ is a regularization
factor. The regularization improves numerical stability and reduces the sensitivity to noise and overfitting. We apply ``weighted ridge-regression'' \cite{ramsay1977comparative,montgomery2012introduction} to train the readout weights to combat the problem of unbalanced cardiac beat classes. To compute this robust ridge estimate, the formula is

\begin{equation}\label{eq5}
	 \mathbf{W}^{out} = \mathbf{Y}_{target}\mathbf{Z}^{T}\mathbf{V}(\mathbf{Z}\mathbf{Z}^{T}\mathbf{V} + \gamma^{2}I)^{-1}
\end{equation}
where the weight matrix, $\mathbf{V} \in R^{(N_{x}+N_{u}) \times (N_{x}+N_{u})}$, is a diagonal matrix with diagonal
elements $V_{ii}$ which are the weights applied to the extended system states and are intended to boost the underrepresented data (i.e., minor heartbeat classes in a single ECG recording). The weights $V_{ii}$ are determined from the extended system states by applying Eq. \ref{eq6} where $\psi(r) = r~exp (- a~|r|)$, $\mathbf{r} =  \mathbf{Y}_{target} - \widetilde{\mathbf{W}}^{out} \textbf{Z}$, $s \geq 1$ and $a \geq 1$ are scaling factors and $\widetilde{\mathbf{W}}^{out}$ is estimated by Eq. \ref{eq4}.

\begin{equation}\label{eq6}
	 \begin{cases}
      V_{ii} = \frac{\psi(r_{i}/s)}{r_{i}/s}& r_{i} \neq 0 \\
      1 & r_{i} = 0. \\
   \end{cases}
\end{equation}

The weight assigned to each residual is controlled by $a$ and $s$ such that the weight is asymptotic to zero for large $|r|$. 

The cardiac beats classification experiment aims to detect pathological conditions such as left bundle branch block (L), right bundle branch block (R), atrial premature beats (A), premature ventricular contraction (PVC), ventricular flutter wave (VF), ventricular escape (VE), fusion of PVC and normal beat as well as paced beats and fusion of paced and normal beats. Figure \ref{beatSample} shows the time course of these target beats compared to normal heartbeats. 

The pathological beat detection task is formalized as a supervised temporal classification problem which demands multiple binary output signals ${\mathbf{Y}_{target}}_{j}(t)$ for the entire time course of heartbeats that belong to $j$-th pathological class:

\begin{equation} \label{eq7}
\resizebox{\hsize}{!}{$
{\mathbf{Y}_{target}}_{j}(t) = \begin{cases}
    1  &  \text{if \textit{t} is in time course of a heartbeat belonged to the $j$-th pathological class},\\
    0  &  \text{otherwise}.
  \end{cases}$}
\end{equation}

Two input signals (i.e., a DC bias set at $1.0$ and the lead II ECG time-series) are fed to the network to stimulate the reservoir neurons; the reservoir responses are then calculated following the Eq. \ref{eq3} and the readout weights are obtained using weighted ridge-regression where the input for linear regression is the smoothed reservoir responses and the target outputs are $\{0,1\}$-valued signals indicating the correct labels of heartbeats.

In Section \ref{Results}, the standard metrics of accuracy (Acc), sensitivity (Se), precision (P), and F1-score are reported to evaluate the classification performance for each subject.

\section{Experimental Results}\label{Results}

The proposed model is grounded on a reservoir of leaky-integrator neurons (Eq. \ref{eq3}) with a few control parameters whose values determine the emergent behavior of the system in response to the input ECG signal. In this section, along with a brief review of the role of each
parameter in the dynamics of the reservoir computing model, the results of customized cardiac beat classification simulations are presented in terms of above-mentioned performance metrics. 

\subsection{Reservoir Parametric Setting}
For producing a functional reservoir for ECG beat classification, it is important to create a memory of the input to provide a temporal context for personalized ECG processing. At the same time, it should be taken into account that a reservoir mainly serves as a nonlinear high-dimensional expansion $\mathbf{x}(t)$ of the input signal $\mathbf{u}(t)$ in a way that in a classification task, the non-linearly separable input data become separable in the extended space of $\mathbf{x}(t)$. Therefore, a rich and relevant enough signal space in $\mathbf{x}(t)$ should be provided such that the desired $\mathbf{Y}_{target}(t)$ could be obtained from it by a linear combination.
Given the model of Eq. \ref{eq3}, the defining parameters of the reservoir of LI neurons are $\mathbf{W^{in}}$, $\mathbf{W}$, and the neuronal leaking rate, $\alpha$. The input and recurrent connection matrices $\mathbf{W^{in}}$ and $\mathbf{W}$ are generated randomly according to adjustable parameters such as scaling(-s) of $\mathbf{W^{in}}$, the reservoir size, $N_{x}$, sparsity, distribution of nonzero elements, and spectral radius of $\mathbf{W}$. In this study, the following choices of parameters were made to fine-tune the reservoir for the task at hand:

\begin{itemize}
  \item Input scaling: to scale the uniformly distributed $\mathbf{W^{in}}$, the input scaling $c$ is defined as the range of the interval $[-c,c]$ from which values of $\mathbf{W^{in}}$ are sampled. In this study, since one channel of ECG recordings (lead II) and a DC signal (the bias) serve as the inputs, a two columns $\mathbf{W^{in}}$ needs to be scaled. The amplitude of ECG wave forms at a given time instance may have a value in $[-3.0,3.0]$ interval, hence, the DC bias was set to $1.0$ which is smaller than the peak-to-peak amplitude of ECG signals. In order to reduce the number of freely adjustable parameters, a single scaling value was then used to scale both columns. 
  \\
  \item Reservoir size: the size of the reservoir is defined as the number of units in the reservoir and determines the dimension of the space of reservoir signals $\mathit{x(n)}$. Intuitively, with a bigger reservoir, it is easier to find a linear combination of the signals to approximate $\mathbf{Y}_{target}(t)$. In our cardiac beat classification task, since the ultimate objective is to implement the reservoir model on available hardware platforms, $N_x = 768$ was opted to make the architecture compatible with the \textit{Dynap-se} processor, a neuromorphic hardware used for implementing analog spiking neural network \cite{moradi2018scalable}. In fact, $N_x$ was set to a surely large enough value and then the model capacity was sought to be confined to the best value by regularization with $\gamma$ (Eq. \ref{eq5}). 
  \\
  
  \item Sparsity, $p$, and the distribution of nonzero elements: the matrix $\mathbf{W}$ is typically generated sparse, with nonzero elements having either a symmetrical uniform or a Gaussian distribution centered around zero. In this study, to speed up the computations, each reservoir node is randomly connected to 10 of other nodes. 
  \\
  \item Spectral radius of $\mathbf{W}$: one of the most central parameters of this model is the spectral radius
of the reservoir connection matrix $\mathbf{W}$, i.e., the maximal absolute eigenvalue of this matrix. This parameter scales the width of the distribution of nonzero elements of $\mathbf{W}$ and is selected to maximize the performance. For the tasks where the current output $\mathbf{y}(t)$ depends more on the recent history of the input, $\mathbf{u}(t)$, smaller values of spectral radius, $\rho(\mathbf{W})$, are typically used. 
\\
\item Leaking rate, $\alpha$: as mentioned above, the leaking rate of the reservoir nodes in Eq. \ref{eq3} can be regarded as the speed of the reservoir update dynamics discretized in time. This parameter is typically set to match the speed of the dynamics of input/output signals; as a small value of $\alpha$ induces slow dynamics of $\mathbf{x}(t)$, for the ECG-oriented tasks, higher values of $\alpha < 1$ is required to capture the slight variations in the time-series.  
\end{itemize}

For the task at hand, in order to optimize the spectral radius and the leakage rate, three ECG recordings were randomly selected (i.e., files \#106, \#119, \#232) and the reservoir was exploited to detect all the annotated QRS complexes regardless of the category to which these cardiac events belong. Within a three-fold cross-validation setup, extensive number of manual experiments were then conducted to adjust the parameters to maximize the QRS detection performance. Table \ref{Table1} summarizes the global parameters defining the designed fixed reservoir.

\begin{table}[h]
\renewcommand{\arraystretch}{1.3}
\caption{Global parameters of the fixed reservoir.}
\label{Table1}
\begin{center}
\resizebox{\hsize}{!}{
\begin{tabular}{|c||c|c|c|c|c|}
\hline
\textbf{Parameter}  & \textbf{input scaling}  & \textbf{reservoir size}  & \textbf{sparsity} & \textbf{spectral radius} & \textbf{leakage rate}\\
\hline
\textbf{Value}  &  0.5 &768    & 0.013\% & 0.99 & 0.99\\
\hline
\end{tabular}
}
\end{center}
\end{table}

Unlike the reservoir parameters which are fixed across all the subjects, the regularization parameter (i.e., $\gamma$ in Eq. \ref{eq5}) and the scaling factors, $a$ and $s$, governing the weighted ridge-regression routine (Eq. \ref{eq6}) are adjusted for every subject through a 5-fold cross-validation approach. 

\begin{figure*}[h!]
\centering
\includegraphics[width= \linewidth]{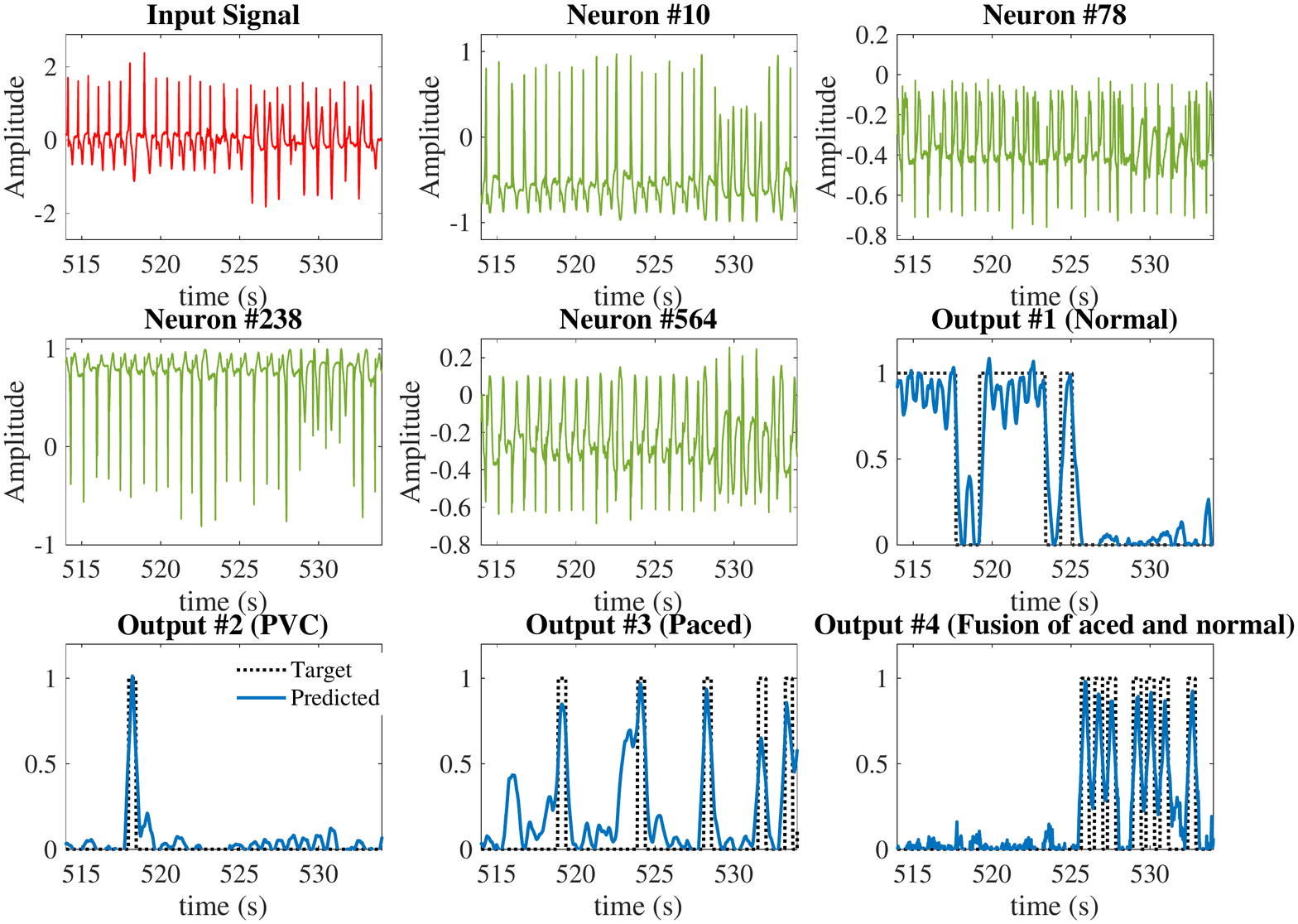}
\caption{A stream of 20 seconds input ECG recording (from file \#217 in MIT-BIH arrhythmia database), activation waveforms of 4 randomly selected reservoir neurons, and activation of 4 output neurons which are trained to detect normal events, PVCs, paced beats and fusion of paced and normal cardiac events.}
\label{fig_activation}
\end{figure*}

\subsection{Performance Evaluation}
The reservoir-based patient-adaptive scheme was applied to the most challenging 15-min time frame of each record in the open-source MIT-BIH arrhythmia database. In contrast to the rest of the signal where pathological cardiac events either occur continuously over a time course of a few minutes or have the same number of events as normal beats, in selected 15-min time frames, the pathological cardiac beats randomly appear in short intervals and have far less number of occurrence than the normal beats (see Table \ref{Table2} for more details). This choice of ECG time course mainly challenges the performance of reservoir computing model in dealing with unbalanced data for early diagnosis applications.

Since the task aimed at classifying short-time events (cardiac beats), the training was set up such that the output $\mathbf{y}(t)$ has a dimension for every pathological condition and $\mathbf{Y}_{target}$ is equal to one for the entire event in the dimension corresponding to that pathological class and zero everywhere else. For each subject, the model is trained to approximate $\mathbf{Y}_{target}$ and the class label at each time instance is decided by 
\begin{equation} \label{eq8}
\begin{split}
\mathbf{Y}_{predicted}(t) &= \operatorname*{arg\,max}_j (\frac{\Delta t}{\tau + \Delta t}\sum_{m = \frac{t - \frac{\tau}{2}}{\Delta t}}^{\frac{t + \frac{\tau}{2}}{\Delta t}} y_{j}(m \Delta t)) \\
& = \operatorname*{arg\,max}_j((\sum{\mathbf{y}})_{j})
\end{split}
\end{equation}
where $y_{j}(t)$ is the $j$th dimension of $\mathbf{y}(t)$ produced by reservoir model from $\mathbf{u}(t)$ and $\tau$ is some integration interval. In this study, $\tau$ was adjusted through the same 5-fold cross-validation approach which has been followed to optimize the regularization parameter, $\gamma$, and weighted ridge-regression scaling factors (i.e., $a$ and $s$). Here, $\sum{\mathbf{y}}$ stands for $y(t)$ time-averaged over a window of length $\tau$ centered at $t$ and is calculated as follows:
\begin{equation} \label{eq9}
\begin{split}
\sum{\mathbf{y}} & = \frac{\Delta t}{\tau + \Delta t}\sum_{m = \frac{t - \frac{\tau}{2}}{\Delta t}}^{\frac{t + \frac{\tau}{2}}{\Delta t}} y(m \Delta t)) \\
&= \frac{\Delta t}{\tau + \Delta t}\sum_{m = \frac{t - \frac{\tau}{2}}{\Delta t}}^{\frac{t + \frac{\tau}{2}}{\Delta t}} \mathbf{W}^{out} [\mathbf{u}(m \Delta t);\mathbf{x}(m \Delta t)] \\
& = \mathbf{W}^{out} \frac{\Delta t}{\tau + \Delta t}\sum_{m = \frac{t - \frac{\tau}{2}}{\Delta t}}^{\frac{t + \frac{\tau}{2}}{\Delta t}}  [\mathbf{u}(m \Delta t);\mathbf{x}(m \Delta t)] \\
&= \mathbf{W}^{out} \sum{\mathbf{z}}
\end{split}
\end{equation}
where $\sum{\mathbf{z}}$ is a shorthand for $[\mathbf{u}(t);\mathbf{x}(t)]$ time-averaged over $\tau$. In another word, to classify the cardiac beats, the model is trained (using Eq. \ref{eq5}) on the time-averaged activations. Once the continuous signal of $\mathbf{Y}_{predicted}(t)$ was calculated following Eq. \ref{eq8}, the label of $i$th QRS complex is defined as the dominant output over the entire cardiac beat. Figure \ref{fig_activation} depicts a short time ECG signal obtained from file \#217 in the MIT-BIH arrhythmia database, samples of the reservoir activations in response to this input as well as the waveforms of three outputs of the trained network. 

The proposed model was individually trained on 7.5-min ECG recordings and tested on the next 7.5-min sequences for every subject. The performance of the models was evaluated using accuracy (Acc), sensitivity (Se), precision (P) and F1-score which are determined by the quantity of true positive (TP), true negative (TN), false negative(FN), false positive (FP) as follows:

\begin{align*}
\begin {split}
Acc &= \frac{TP + TN}{TP + FP + TN + FN},\\
Se &= \frac{TP}{TP + FN},\\
P &= \frac{TP}{TP + FP},\\
F1-score &= \frac{2TP}{2TP + FP + FN}.
\end{split}
\end{align*} 

In cases where a patient exhibited several abnormality classes, separate reports on different classes were provided.  
Table \ref{Table2} presents the performance of the proposed reservoir computing model on 17 records. More details on patient specific control parameters are also reported in the same table.

As it can be seen from Table \ref{Table2}, the proposed algorithm exhibited excellent performance in terms of personalized accuracy, sensitivity, precision, and F1-score. However, the global metrics, when calculated as the non-weighted average of personalized metrics over all the subjects and classes, are highly affected by the relatively poor performance of the algorithm on subjects identified by the files \#209 and \#207 in the dataset. In previous works with emphasis on subject-based assessments (e.g., \cite{de2004automatic,engin2004ecg,faezipour2010patient,dutta2010correlation,chen2017heartbeat}), neither of these records were explored. Excluding these two files, a global accuracy of 0.9911 obtained in current experiments excels the values reported in  \cite{chen2017heartbeat,engin2004ecg,dutta2010correlation,de2004automatic} which are 0.9846, 0.967, 0.9582, 0.936, respectively. The global sensitivity and precision of the proposed method (excluding those challenging files) are 0.9638 and 0.9725 and outperform the values reported in studies \cite{de2004automatic,dutta2010correlation,chen2017heartbeat} where subject-specific cardiac beat classification was of interest.
It is also worth mentioning that the proposed method detects premature ventricular contractions (PVCs) with a global accuracy of 0.9929 and specificity of 0.9868. As a PVC detector, it can be perceived in Table \ref{Table2} that the performance of the model is worse in ECG-records associated with file \#213 where there are only 53 PVCs among 814 tested cardiac beats. Excluding this subject, the proposed model outperforms existing PVC detectors in terms of sensitivity and precision as well as accuracy \cite{dutta2010correlation,zhang2014heartbeat, chen2017heartbeat}.

Taking all these reported measures into consideration, however, since the performance of many algorithms for automatic heartbeats classification were obtained after training and testing on different choices of ECG recordings, various time slots and diverse numbers of normal and pathological cardiac beats, a comparison is difficult and a comparative study needs to be conducted to compare the classification abilities of the proposed method with those of previously developed algorithms. It can only be concluded that the current hardware-compatible proposed approach is able to detect pathological beats with high accuracy even when only a few samples of a target class are presented.

\begin{landscape} 
\fontsize{9}{9}\selectfont
\renewcommand{\arraystretch}{1.5}
\centering
\begin{longtable}{|>{\centering\arraybackslash}p{1.5cm}
                ||>{\centering\arraybackslash}p{4.0cm}
                |>{\centering\arraybackslash}p{5.0cm}
                |>{\centering\arraybackslash}p{5.0cm}|} 
\caption{Patient-specific control parameters and performance evaluation of the reservoir computing model on MIT-BIH arrhythmia database.} \label{Table2}\\

\hline \multicolumn{1}{|>{\centering\arraybackslash}p{1.5cm}||} {\textbf{Data \#}} & \textbf{Pathological Classes} & [\textbf{Test time slot, \#Beats, Regularization parameter, Ratio of scaling factors} ($\frac{a}{s}$), \textbf{$\tau$]}   & [\textbf{Acc}  \textbf{Se} \textbf{P} \textbf{F1-score}] \\ \hline 
\endfirsthead
\multicolumn{4}{r}{\footnotesize Continue on the next page}
\endfoot
\endlastfoot
 \multirow {2}{*}{'104'} & {Class I: Paced beats (\#~beats = 380)} &  [[22.5,30] min, 554, 1e-5, no weights, 0.91] & {Class I: [0.9928    0.9947    0.9947    0.9947]} \\
  &  {Class II: Fusion of paced and normal beats (\#~beats = 151)} &  [[22.5,30] min, 554, 1e-6, 3.0, 0.83] & {Class II: [0.9856    0.9934    0.9554    0.9740]} \\ 
  \hline
  {'106'} & {Class I: Premature ventricular contraction (\#~beats= 148)} & [ [17.5,25] min, 501, 0.001, 1.5, 0.52] & {Class I: [0.9940    0.9932    0.9866    0.9899]} \\
  \hline
 {'119'} & {Class I: Premature ventricular contraction (\#~beats= 156)} & [[17.5,25] min, 499, 1e-5, no weights, 1.0] & {Class I: [0.9980    1.0000    0.9936    0.9968]} \\
  \hline
  {'200'} & {Class I: Premature ventricular contraction (\#~beats= 205)} & [[12.5,20] min, 663, 0.01, no weights, 0.5] & {Class I: [0.9910    0.9951    0.9761    0.9855]} \\
  \hline
 {'201'} & {Class I: Premature ventricular contraction (\#~beats= 103)} & [[11.5,19] min, 396, 1e-7, no weights, 1.0] & {Class I: [0.9924    1.0000    0.9717    0.9856]} \\
  \hline
  {'203'} & {Class I: Premature ventricular contraction (\#~beats= 84)} & [[12.5,20] min, 483, 1e-6, no weights, 0.02] & {Class I: [0.9814    1.0000    0.9032    0.9492]} \\
  \hline
  \multirow{5}{*}{'207'} & {Class I: Left bundle branch block beat (\#~beats= 523)} & [[7.5,15] min, 523, 1e-6, no weights, 1.22] & {Class I: [1.0     1.0     1.0     1.0]} \\
  &  {Class II: Atrial premature beat (\#~beats= 65)} & [[29,30] min, 65, 1e-6, no weights, 1.22] & {Class II: [1.0     1.0     1.0     1.0]} \\
  &  {Class III: Premature ventricular contraction (\#~beats= 51)} & [[2.5,5] min, 218, 1e-6, 1.5, 1.22] & {Class III: [0.9404    0.9608    0.8167    0.8829]} \\
  &  {Class IV: Ventricular flutter wave (\#~beats= 187)} & [[26.5,29] min, 283,1e-7, no weights, 1.5] & {Class IV: [0.9894    0.9840    1.0000    0.9919]} \\
  &  {Class V: Ventricular escape beat (\#~beats= 68)} & [[28,30] min, 178, 0.1, no weights, 1.5] & {Class V: [0.9944    0.9853    1.0000    0.9926]} \\
  \hline
  \multirow{2}{*}{'208'} & {Class I: Premature ventricular contraction (\#~beats= 303)} & [[7.5,15] min, 728, 1e-6, no weights, 0.64] & {Class I: [0.9918    0.9934    0.9869    0.9901]} \\
  & {Class II: Fusion of ventricular and normal beats (\#~beats= 149)} & [7.5,15] min, 728, 1e-5, 2.0, 0.78] & {Class II: [0.9959    0.9933    0.9867    0.9900]} \\
  \hline
 {'209'} & {Class I: Atrial premature beat (\#~beats= 165)} & [[10,17.5] min, 774, 1e-3, 2.5, 0.92] &{Class I: [0.9483    0.8485    0.9032    0.8750]} \\
  \hline
 {'212'} & {Class I: Right bundle branch block beat (\#~beats= 519)} & [[7.5,15] min,702, 1e-6, no weights, 0.92] & {Class I: [1.0  1.0  1.0  1.0]} \\
  \hline
  \multirow{2}{*}{'213'} & {Class I: Premature ventricular contraction (\#~beats= 53)} & [[9,16.5] min, 814, 1e-4, 2.5, 0.78] & {Class I: [0.9840    0.9245    0.8448    0.8829]} \\
  & {Class II: Fusion of ventricular and normal beat (\#~beats= 78)} & [[[9,16.5] min, 814, 1e-6, 1.5, 1.0] ]& {Class II: [0.9668    0.8200    0.8200    0.8200]} \\
  \hline
  \multirow{3}{*}{'217'}& {Class I: Premature ventricular contraction (\#~beats= 30)} & [[21.5,29] min, 543, 1e-5, 2.5, 0.72] & {Class I: [1.0  1.0  1.0  1.0]} \\
  & {Class II: Paced beat (\#~beats= 31)} & [[21.5,29] min, 543, 1e-5, 2.5, 1.0] & {Class II: [0.9908    0.9706    0.8919    0.9296]} \\
  & {Class III: Fusion of paced and normal beat (\#~beats= 454)} & [[21.5,29] min, 543, 1e-5, no weights, 0.78] & {Class III: [0.9945    0.9978    0.9956    0.9967]} \\
  \hline
  {'221'} & {Class I: Premature ventricular contraction (\#~beats= 50)} & [[21.5,29] min, 569, 1e-7, 1.5, 0.78] & {Class I: [1.0  1.0  1.0  1.0]} \\
  \hline
 {'223'} & {Class I: Premature ventricular contraction (\#~beats= 203)} &  [[21.5,29] min, 649, 0.8e-5, 1.3, 0.56] & {Class I: [0.9908    0.9803    0.9900    0.9851]} \\
  \hline
 {'231'} & {Class I: Right bundle branch block beat (\#~beats= 269)} & [[7.5,15] min, 507, 1e-4, no weights, 0.92] & {Class I: [0.9980    1.0000    0.9963    0.9981]} \\
  \hline
 {'232'} & {Class I: Atrial premature beat (\#~beats= 366)} & [[7.5,15] min, 465, 1e-4, 1.5, 1.5] & {Class I: [0.9763    0.9837    0.9863    0.9850]} \\
  \hline
 {'233'} & {Class I: Premature ventricular contraction (\#~beats= 207)} & [[16.5,24] min, 770, 0.001, no weights, 0.17] & {Class I: [0.9987    1.0000    0.9952    0.9976]} \\
  \hline
  
\end{longtable}
\end{landscape}

\section{Conclusion}\label{Conclusion}

Investigating the possibility of applying a hardware-compatible variation of reservoir models to the heartbeat classification problem, an echo state network was trained and evaluated on several benchmark signals from the MIT-BIH arrhythmia database. It was shown that the proposed model is able to efficiently detect pathological cardiac beats with high accuracy even when only a few samples of a target class are presented. The overall accuracy and F1-score were found to be as encouraging as 99\% and 97\%, respectively. The F1-score was reported as the harmonic average of the precision and sensitivity to evaluate the performance of the scheme for subject-based binary (normal vs pathological) classification problems.\\ 

The primary objective of this research was to conduct a complete set of experiments to verify the reliability of this model for cardiac beat monitoring application. This model can be implemented on digital systems such as field programmable gate arrays (FPGAs) and application specific integrated circuit (ASIC) chips in both embedded and stand-alone systems. Our proposed model, for instance, is compatible with current FPGA-based ESN architectures which offer leaky neuegrator ronintal units, a neuronal arithmetic section, input/output encoding/decoding components, memory control, memory blocks such as RAMs and/or memristors wherein the neuronal and synaptic weights is stored, and a Read/Write interface to access to these memory blocks \cite{schrauwen2007compact}. On analog electronic systems, however, realizing a functioning reservoir entails tackling a number of critical issues related to input encoding, memory organization, parallel processing, on-chip learning and tradeoffs between area, precision and power overheads. As a further extension to this study, therefore, the ``reservoir transfer learning method'' (developed in \cite{he2018EMBS}) will be applied to implement this model on \textit{Dynap-se} analog neuromorphic microprocessor \cite{moradi2018scalable}.

\section*{Acknowledgment}
This work was supported by European H2020 collaborative project NeuRAM3 [grant number 687299]. I would also like to thank Herbert Jaeger, who provided insight and expertise that greatly assisted this research. 


\bibliography{mybibfile}

\end{document}